\documentclass[letterpaper, 10 pt, conference]{ieeeconf} 
\IEEEoverridecommandlockouts

\usepackage{cite}
\usepackage{amsmath,amssymb,amsfonts}
\usepackage{algorithmic}
\usepackage{graphicx}
\usepackage{textcomp}
\usepackage{xcolor}

\usepackage[caption=false, font=footnotesize]{subfig}
\usepackage{threeparttable}
\usepackage{booktabs}
\usepackage{multirow}

\PassOptionsToPackage{hyphens}{url}
\usepackage[pagebackref,breaklinks,colorlinks]{hyperref}
\usepackage{cleveref}

\begin{document}

\title{ViewpointDepth: A New Dataset for Monocular Depth Estimation Under Viewpoint Shifts}

\author{Aurel Pjetri$^{1,2\dagger}$, Stefano Caprasecca$^{1}$, Leonardo Taccari$^{1}$, Matteo Simoncini$^{1}$, Henrique Piñeiro Monteagudo$^{1,3}$,\\Wallace Walter$^{5*}$, Douglas Coimbra de Andrade$^{4*}$, Francesco Sambo$^{1}$ and Andrew David Bagdanov$^{2}$\\
\small\texttt{{\href{https://www.kaggle.com/datasets/aurelpjetri/viewpointdepth}{kaggle.com/datasets/aurelpjetri/viewpointdepth}}}
\thanks{$^{1}$Verizon Connect, Florence, Italy. $^{2}$University of Florence, Florence, Italy. $^{3}$University of Bologna, Bologna, Italy. $^{4}$SENAI Institute of Innovation, Rio de Janeiro, Brazil. $^{5}$Retired. $^{*}$Worked on this paper while at Verizon Connect.}
\thanks{$^{\dagger}$Email: \texttt{\href{mailto:aurel.pjetri@verizonconnect.com}{aurel.pjetri@verizonconnect.com}}}
\thanks{H.P.M. acknowledges support from the SMARTHEP project, funded by the European Union’s Horizon 2020 research and innovation programme, call H2020-MSCA-ITN-2020, under Grant Agreement n. 956086.}
}

\maketitle

\begin{abstract}
Monocular depth estimation is a critical task for autonomous driving and many other computer vision applications. While significant progress has been made in this field, the effects of viewpoint shifts on depth estimation models remain largely underexplored. This paper introduces a novel dataset and evaluation methodology to quantify the impact of different camera positions and orientations on monocular depth estimation performance. We propose a ground truth strategy based on homography estimation and object detection, eliminating the need for expensive LIDAR sensors. We collect a diverse dataset of road scenes from multiple viewpoints and use it to assess the robustness of a modern depth estimation model to geometric shifts. 
After assessing the validity of our strategy on a public dataset, we provide valuable insights into the limitations of current models and highlight the importance of considering viewpoint variations in real-world applications.
\end{abstract}

\section{Introduction}
\label{sec:intro}
Autonomous driving has been receiving more and more attention from the public with different forms of it being regulated and approved for public implementations, from fully autonomous robotaxis \cite{waymo} to consumer vehicles with different levels of autonomy \cite{mercedes,tesla}. 
Perception, and computer vision in particular, is the most important building block for these technologies. A fundamental task for safety and perception of surroundings is depth estimation. Despite the recent advancements in machine learning, it is well known that such tasks are susceptible to out-of-distribution samples, from adverse weather and light conditions to image corruptions~\cite{Saunders_2023_ICCV, Wang_2021_ICCV, kong2023}.

Viewpoint change is another important distribution shift for depth estimation,
either caused by vehicles with different sizes or by variations in the camera installation. 
A few works have only recently started to shed light on the effect of such shifts on related tasks like Bird's Eye View (BEV) semantic segmentation and object detection~\cite{klinghoffer2023, vidit2023}. 
However, quantification of such an effect on monocular depth estimators remains heavily underexplored, with only \cite{dijk2019} giving some insights on the behaviours of these models.
This could be mainly due to the lack of annotated data and the cost and technical difficulties associated with it. Traditionally, this kind of task is performed using measurements from expensive LIDAR sensors~\cite{geiger2013, yogamani2019, caesar2020}. These sensors are not ubiquitous and require multiple layers of post-processing, such as outlier detection and removal, data alignment, registration with other sources, and other computationally expensive steps. A possible solution lies in the domain of synthetic data, which introduces a well-known sim-to-real domain shift~\cite{hu2022}.

\begin{figure}[tb]
\centering
\subfloat[]{\includegraphics[width=0.5\linewidth]{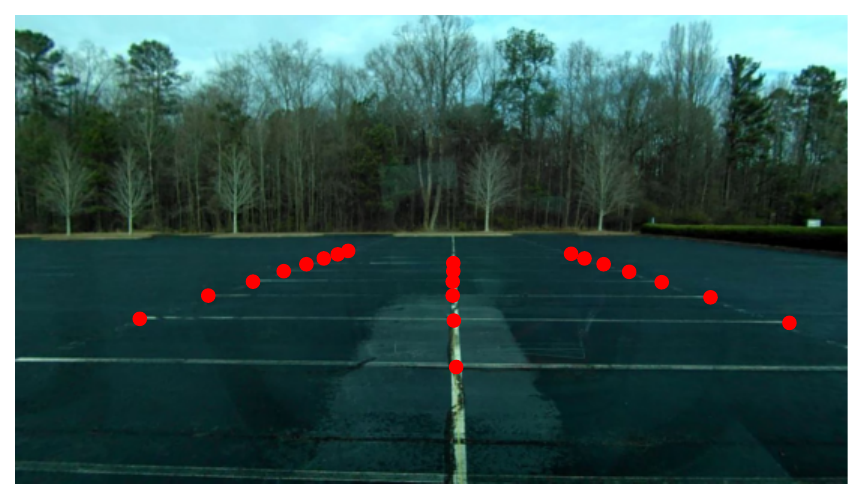}
\label{fig:hom_a}}
\subfloat[]{\includegraphics[width=0.4\linewidth]{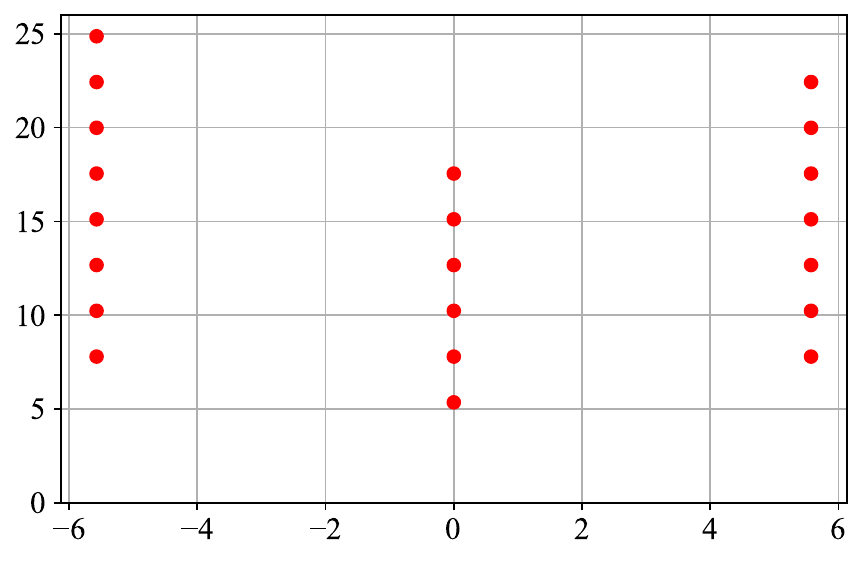}
\label{fig:hom_b}}
\caption[]{\subref{fig:hom_a} Manually labeled source points for the homography. \subref{fig:hom_b} 
Target points for the homography with metric distances.
}
\label{fig:hom}
\end{figure}

Our proposal is to remove the dependency on expensive sensors to evaluate depth estimation techniques by exploiting basic geometric elements such as \emph{homographies}.
We estimate a ground truth (GT) homography from an initial calibration session that maps points from the image plane into the road plane (see Fig. \ref{fig:hom}). Then, we use the homography to compute distances of road objects that serve as GT for our evaluation.
Our proposal is to use \emph{object distance estimation} as a proxy task for traditional depth evaluation.
Although this approach is accurate only if the geometry of the road plane is in line with the one measured during calibration, we empirically show that estimation error is considerably limited on planar road scenes such as those in the KITTI dataset~\cite{geiger2013}. Furthermore, experiments on KITTI show that our proxy task provides evaluations that are very close those observed using LIDAR GT.

Our strategy enabled us to collect a large dataset of video sequences, recorded from different known and calibrated camera settings. We use such data to highlight the impact of the different viewpoint shifts on a modern depth estimator and to give insights into their effects on the scale perceived by the model.

To summarize, the main contributions of this work are:
\begin{itemize}
    \item we propose a new ground truth strategy based on classical computer vision techniques, which can be used to replace expensive sensors for object distance estimation;
    \item we discuss in depth the limitations of the proposed approach and empirically show its effectiveness on a public annotated dataset;
    \item we release a new dataset of road scenes collected from multiple viewpoints; and
    \item we quantify the effects of different viewpoints on the performance of a modern monocular depth estimation model.
\end{itemize}

\section{Related Work}
\label{sec:related}
\subsection{Road Scene Depth Datasets}
\label{subsec:related-dataset}
Many datasets have been proposed to facilitate research on 3D perception for autonomous driving or, in general, for road scene analysis. The KITTI dataset was a pioneering work in this context, in which a test vehicle collected driving data from multiple cameras and sensors, including a laser scanner for distance ground truth~\cite{geiger2013}. The authors collected different scenes driving in an urban environment in Germany.
Inspired by KITTI, other datasets have been released by expanding the variety of geographical regions and scenes collected. Examples are Nuscenes~\cite{caesar2020} and Waymo Open~\cite{sun2020}, which are larger than KITTI and cover a wider variety of cities and scenes.
Other datasets like WoodScape~\cite{yogamani2019} focus on fisheye cameras, which are not only front-facing, but also cover lateral and rear, and are mounted on many modern cars.

All public datasets, in general, are mainly focused on capturing a variety of scenes, weather, lighting conditions, sensors or image corruptions~\cite{kong2023}. None of them, however, provides a broad variety of viewpoints, as the vehicle type and camera position are usually fixed across the whole dataset.

\subsection{Self-supervised Monocular Depth Estimation}
\label{subsec:related-depth}
Depth estimation is a fundamental geometric task for 3D scene understanding. Self-supervised monocular depth estimation is a particular variant of this task that enables training networks without the use of explicit depth labels. Although it is an ill-posed problem, since multiple 3D scenes can generate the same 2D image, some deep learning techniques have shown good results. One of the first proposed approaches used the pretext task of reconstructing frames in a video sequence to force the network to learn a good depth estimation of the scene~\cite{zhou2017}. The depth network is trained alongside a pose network for the new frame to be rendered. Given a source frame at time $t$ and a target frame at $t+1$, each pixel $p_{t+1}$ can be back-projected to the source view using depth $D_t$ and pose $P_{t+1 \rightarrow t}$ predicted by the networks: $p_t \sim KP_{t+1 \rightarrow t}K^{-1}p_{t+1}$.
In this way bilinear sampling can be used to synthesise $\hat{p}_{t+1}$ \cite{jaderberg2015}, and a direct comparison of each pixel within each frame can then be used for training loss~\cite{zhou2017}.
Inspired by this work,~\cite{godard2019} introduced multiple improvements, such as the use of a photometric error in the loss function combining L1 and SSIM~\cite{zhou2004}. The results were impressive, not only because they made state-of-the-art, but also given how close they were to stereo-based methods, proving the effectiveness of this paradigm.

Over the years many works have built on top of this paradigm, implementing more effective and modern architectures. One of the first implementations of vision transformers~\cite{dosovitskiy2021} to tackle this problem was proposed in~\cite{zhao2022}. In particular, inspired by the MPViT architecture \cite{Lee_2022_CVPR}, the authors propose a Conv-stem block for the network encoder, in which they combine convolutional blocks with transformers. They observe that this enables modeling both local and global information in the frame, achieving state-of-the-art at the time of its release.

\subsection{Viewpoint Shift}
\label{subsec:related-shift}
Depth estimation under geometric shifts remains underexplored, despite being a well-known problem.
It has been tackled in the context of supervised depth estimation by introducing a geometric embedding of the inputs and a variational latent representation~\cite{Guizilini_2023_ICCV}.
Additionally, viewpoint shifts caused by different camera installations for autonomous driving have been studied~\cite{klinghoffer2023}. In this case, the task was bird's eye view semantic segmentation and the dataset used was synthetic. Moreover, a viewpoint generalization technique for object detection, based on learning homography transformations of the input images to account for different geometric shifts, has been proposed~\cite{vidit2023}.

Finally, for the specific task of monocular depth estimation, \cite{dijk2019} conducted multiple experiments with the aim of understanding which depth cues are actually exploited by the models. 
The authors observed that vertical position of the object is the main cue used by the models: the lower side of the vehicle and the point where it touches the ground seem to be the main cues exploited by models to estimate its distance.
Their experiments also suggest limitations of such models to account for camera orientation changes simulated by image central crops and rotations.

\section{Ground Truth and Evaluation Strategy}
\label{sec:method}
We propose to use a homography in combination with an object detector as a ground truth source for measuring monocular depth estimation error on road scenes. Depth estimation models are traditionally tested by directly comparing per-pixel predictions with ground truth from expensive sensors, such as LIDAR, and computing metrics such as absolute relative error:
\begin{equation}
  \text{abs-rel} = 100\cdot\frac{1}{N} \sum^{N}_{i=0}\frac{\left|x^{i}_{gt}-x^{i}_{pred}\right|}{x^{i}_{gt}}
  \label{eq:err}
\end{equation}

We hypothesize that comparable results can be achieved by evaluating the models on object distance estimation, measuring the error only on road objects instead of on the whole frame. Moreover, we observe that a homography that maps ground points in the image to a plane with a metric reference system can be used as distance ground truth source for points on the ground (Fig. \ref{fig:hom}). Such projection, used in combination with an object detector, can associate a metric distance to objects.
As a matter of fact, previous studies have already suggested that monocular depth models use the lower part of the object to determine their distance \cite{dijk2019}, which further supports our strategy.
This ultimately enables experiments and research in conditions in which expensive sensors are not available. We divide our method in three steps: calibration, object detection, and evaluation.

\subsection{Calibration}
The calibration phase is the only manual step of the method. We select reference ground points and measure their distance to the camera to construct the target plane with a metric reference system centered in the camera (see Fig. \ref{fig:hom}), so that for any point $\text{X}=[x,y]^T$ its distance to the camera can be computed by simply $\sqrt{x^2+y^2+h^2}$, where $h$ is the height of the camera. 
This allows us to estimate a homography mapping points in the image to points on the target plane, and therefore to associate a metric distance to any point in the image, provided that it lies on the ground.

\subsection{Object Detection}
\begin{figure}[tb]
\centering
\subfloat[]{\includegraphics[width=0.5\linewidth]{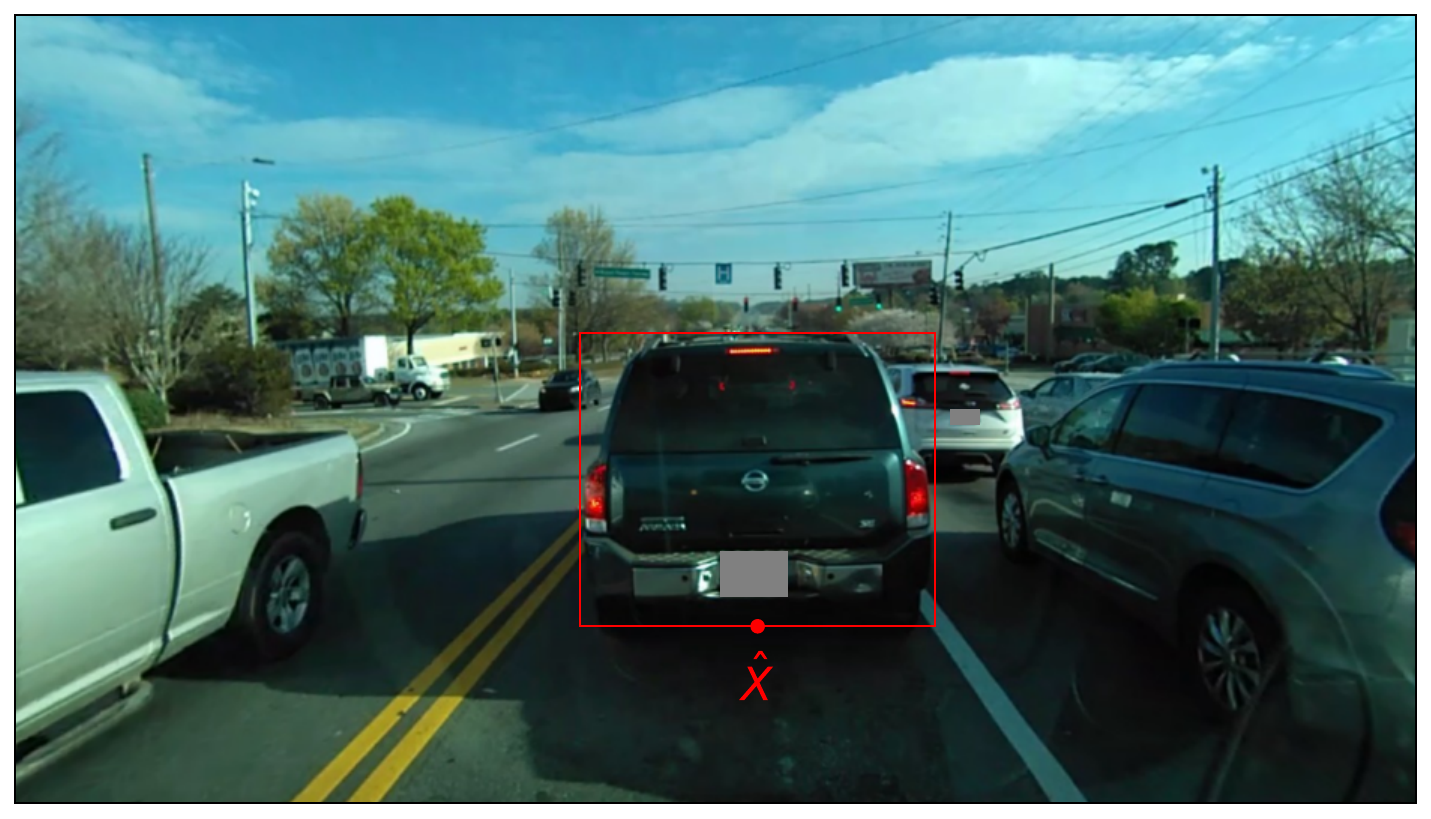}
\label{fig:box_a}}
\subfloat[]{\includegraphics[width=0.5\linewidth]{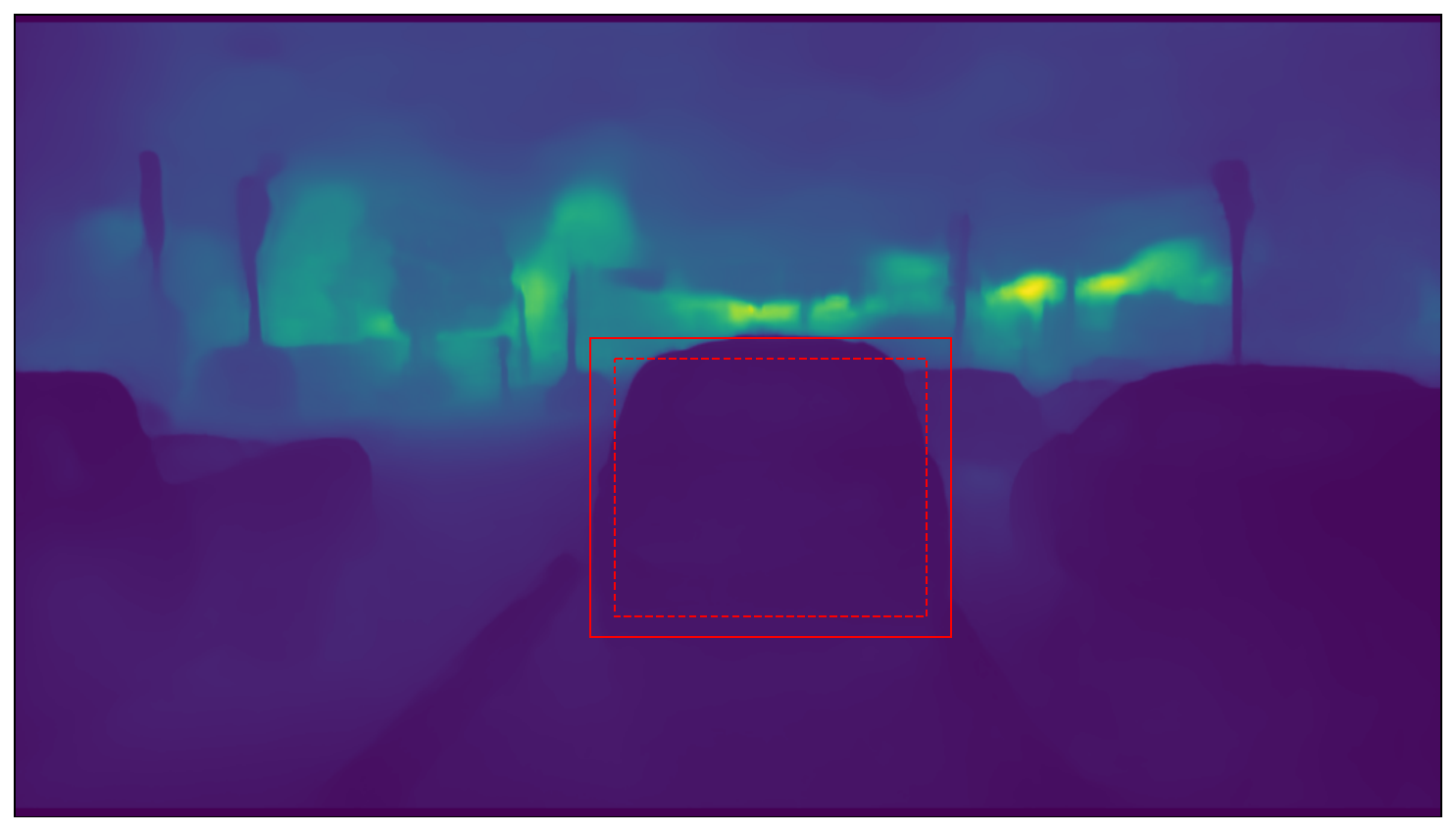}
\label{fig:box_b}}
\caption[]{\subref{fig:box_a} Bounding box of a vehicle with the point $\hat{X}$ used for GT distance. \subref{fig:box_b} Depth prediction. The original box was resized to $\alpha=75\%$ to extract the percentile $\beta$ of the prediction.}
\label{fig:box}
\end{figure}

We use an object detector to detect the relevant road objects in the image: car, truck, bus, motorcycle, bicycle and person.
Given a bounding box with coordinates $(x_1, y_1)$ and $(x_2, y_2)$, we assume that all objects lie on the ground, and therefore we define their distance as that of the center of the lower side of the box $\hat{X} = ((x_2-x_1)/2, y_2)$ (Fig. \ref{fig:box_a}).
By projecting this point we are able to associate a metric distance to each object and use it as ground truth for the model evaluation in the next step.

\subsection{Evaluation}
To measure the performance of a depth estimation model, we only consider road objects. Therefore, we again use the bounding box information from the object detector and 
use the depth estimation model to infer the dense distance of the points within the boxes. 
We then select a percentile $\beta$ from the estimated distances and use that as the inferred object distance. 
Moreover, to account for inaccuracies we resize bounding boxes to a certain fraction $\alpha$ of the original dimension before collecting inferred depths, as shown in Fig. \ref{fig:box_b}. $\alpha$ and $\beta$ are treated as hyperparameters that we select in the experimental phase (see Section \ref{subsec:kitti}).

\section{Dataset}
\label{sec:dataset}
The aim of this dataset is to measure the effects of different viewpoint shifts on monocular depth estimation models. 
We collected the data using a 2023 RAM Promaster 2500 cargo van located in Alpharetta (Georgia), United States. Between March 2023 and February 2024, we installed two identical dashcams in the windshield of the vehicle. The first camera, to which we refer as \emph{base}, was fixed in an ideal position for a dashcam, i.e., close to the rear-view mirror at a height of 177.8~cm with an orientation of 0° roll, 0° yaw and -4° pitch (later moved to 6° pitch as shown in Table \ref{tab:posrot}).

The second camera, which we identify as \emph{shifted}, was instead used to simulate different viewpoint shifts and was moved on a weekly basis. Table \ref{tab:posrot} shows the different combinations of positions and orientations of the cameras during the collection period. For the camera position in the windshield (X, Y, Z) we centered the reference system on the \emph{base} camera on the inside of the car looking forward in the direction of movement, with the X axis oriented horizontally and positive on the right, the Y axis vertically and positive downward and the Z axis oriented in the direction of movement. Moreover roll pitch and yaw are oriented following the right-hand rule.
Roll angles were computed by manually measuring the camera inclination on the windshield. 
On the other hand, we measured pitch and yaw angles for each viewpoint by manually labeling the horizontal vanishing point from the straight lines on the road $(v_u, v_v)$, and computing the angle given the focal length $(f_x, f_y)$ of the camera and its principal point $(c_u, c_v)$, e.g. $(180/\pi)arctan((c_v-v_v)/f_y)$ for pitch.

 \begin{table}[tb]
 \begin{threeparttable}[t]
   \caption{
     Different positions and orientations collected 
   }
   \label{tab:posrot}
   \centering
   \setlength{\tabcolsep}{5pt}
   \begin{tabular}{@{} r rrr rrr r | r r @{}}
     \toprule
 &\multicolumn{7}{c|}{\emph{Shifted} Dashcam}& \multicolumn{2}{c}{\emph{Base} Dashcam}\\
 ID&X&Y&Z&Pitch&Yaw&Roll&Objects&Pitch&Objects\\ 
 \midrule
 0&0&30&0&-4&0.0&0&5291&-4&3026\\
 \midrule
 1&5&0&0&-10&0.0&0&4325&6&4146\\ 
 2&5&0&0&16&0.0&0&5004&6&5458\\ 
 3&5&0&0&-4&2.5&0&2906&6&2824\\ 
 4&5&0&0&-4&5.0&0&2281&6&2219\\ 
 5&5&0&0&16&5.0&0&3068&6&4207\\ 
 \midrule
 6&-65&-5&-10&5&0.0&-5&4425&6&3916\\ 
 7&-65&-5&-10&16&0.0&-5&3537&6&3759\\ 
 \midrule
 8&-65&30&-10&5&0.0&-5&3465&6&3278\\ 
 9&-65&30&-10&16&0.0&-5&3712&6&4413\\  
   \bottomrule
   \end{tabular}
 \begin{tablenotes}
 \item X, Y and Z are expressed in centimeters. Pitch, roll and yaw orientations are in degrees. Objects refers to the number of objects detected.
   \end{tablenotes}
   \end{threeparttable}
 \end{table}

A total of 10 different viewpoints were collected. For each viewpoint we release 3 videos of 10 minutes each from different scenes. Given the long time range in which the videos are distributed, we manually selected samples to represent a similar distribution of suburban areas in non-adverse weather and at day-time. Dashcams recorded at an average of 30 frames per second in 720p MP4 format and with a 150° diagonal field of view. We down-sampled the data at 10 fps and extracted the frames undistorted in JPEG format, with a quality factor of 92 and 4:2:0 chroma sub-sampling. This resulted in a total of 360K frames. 
Note that although the two cameras are simultaneously recording the same scene, no synchronization mechanism was implemented. This can result in the two recordings being shifted by a variable number of frames. 

Finally, to foster further research, we also release GPS points and accelerometer data at 1~and~100~Hz, respectively.

\subsection{Object Detection}
\label{subsec:od}
We used a Yolov5 object detector \cite{yolov5}, trained on a combination of BDD100K \cite{bdd100k} and a private dataset collected with same camera model of this work.
We do not expect the object detector to be affected by the viewpoint shifts as both datasets used for training have non-fixed viewpoints. Moreover, from a qualitative analysis, we observe that the combinations in Table \ref{tab:posrot} do not induce a significant change in appearance of the objects.
This is also confirmed by smaller-scale experiments that showed no evident change in mean average precision between the different viewpoints.

To exclude problematic cases, we filter out all the boxes with a confidence level lower than 0.5 and the lower side above the vanishing point of the camera. 
Furthermore, we remove occluded objects by detecting overlapping boxes and only keeping the lowest ones in the image. Finally, we filter out small objects with specific area thresholds for each class: 3000 pixels$^2$ for all vehicles, 1000 pixels$^2$ for person and bicycle, and 1500 pixels$^2$ for motorcycle.
Both raw and filtered detections are released with the dataset.

\subsection{Homography}
\label{subsec:hom}
As mentioned in Section \ref{sec:method}, in order to be able to associate a ground truth distance to road objects, we estimated homographies mapping image ground points to a target plane with metric reference system (Fig. \ref{fig:hom}) that we release alongside the dataset.

\subsubsection{Base Camera Homography}
We exploited the regular pattern of an empty parking lot in order to have several reference points and minimize manual measurements. 
In Fig. \ref{fig:hom} we show the ground points manually labeled on the image and the corresponding points on a plane with metric coordinates. The homography was estimated using all the provided points with a simple least-squares scheme from the OpenCV \cite{opencv} Python library.

\subsubsection{Shifted Camera Homography}
We selected a set of corresponding points between the cameras after manually synchronizing a portion of the two videos. Using the \emph{base} camera homography, we projected the points from the base camera to the target plane and used the corresponding points from the \emph{shifted} camera to estimate the needed homography. This operation was repeated for each position.

\section{Experimental Results}
\label{sec:exp}
In this section we report the results of our experiments. In particular, we validate our method as ground truth source by comparing it with LIDAR readings on the KITTI dataset in Section \ref{subsec:kitti}. Then, in Section \ref{subsec:exp}, we use our ground truth method to measure the effects of the different viewpoints on a self-supervised monocular depth estimator.

In particular, for all our experiments we use the MonoViT depth estimation model~\cite{zhao2022}, which is a modern architecture which yields state-of-the-art performance on the RoboDepth challenge on robustness against adverse weather, light, and sensor corruption~\cite{kong2023}. Despite being a challenge on simulated weather conditions and image degradation, we thought it would be a promising candidate for our study.

\subsection{Homography Evaluation on KITTI}
\label{subsec:kitti}
The goal of this experiment is to assess the margin of error of our geometry-based strategy by comparing it with LIDAR ground truth. We take the MonoViT model trained on the KITTI dataset \cite{geiger2013} for self-supervised monocular depth estimation and measure its absolute relative error (\ref{eq:err}) using two ground truth sources: laser scans and our strategy.

As a first step we estimate the homography by exploiting the provided camera projection matrix. We select a set of reference points in the target plane with metric reference system, project them to the image plane and estimate the homography with the same method used in our data.
We also exploit the provided object labels \cite{Geiger2012CVPR} and filter out occluded and partial objects.

We perform inference with MonoViT on the test set from the Eigen split \cite{eigen2014}. Monocular depth estimators give predictions with an unknown scale factor and need to be manually scaled to metric depth using ground truth. Traditionally, a median scaling is performed at an image level, i.e. for each image $j$ predictions are multiplied by $s_{j}=\mbox{median}(X^{j}_{gt})/\mbox{median}(X^{j}_{pred})$. However, this can be unfair towards other stereo-based or supervised methods that do not use ground truth information at test time, as highlighted by~\cite{godard2019}. Therefore, in these experiments we compute a single scale multiplier by computing medians across the whole test set. We replicate the same in our strategy by computing medians of homography GTs and prediction percentiles extracted from the boxes.

To find the best parameters for bounding box resize and percentile of the prediction inside the box ($\alpha$ and $\beta$ respectively), we performed a grid search over the two parameters and found that the best performance is given by $\alpha=0.75$ and $\beta=75$. With these parameters we measured an abs-rel of 16.8 with our homography and 13.6 with laser, giving a difference of 3.22 points.

Moreover, for the task of object distance estimation, we compute Spearman rank correlation scores between laser readings and the homography. To associate a laser distance to an object, we take the sensor reading of the middle point of the lower side of the box. What we observe is that the two sources of ground truth have a high correlation with a Spearman score of 0.97.

These results also suggest that the intrinsic error introduced by the use of a homography, with its strong assumptions of flat ground surface, is limited. To further verify this, we also computed statistics about road inclination in KITTI's test set exploiting the provided GPS signals. As a matter of fact, using altitude and speed, we are able to compute the road's slope in the traveled segment. In particular we sample GPS points at 1Hz and keep only those where the horizontal displacement resulted in more that 1m. We observe that the average of the absolute values of the angles is 1.2°, median is 0.4° and 99th percentile is 21.1°. The same statistics on our dataset are: 1.8°, 0°, 19.1° respectively. 
Moreover, the homography's assumption is broken only in the case where there is a change in the road geometry, therefore if we compute the number of times the GPS altitude changes of more than 1m (the granularity of our GPS signal), we observe that in KITTI it happens in 1.46\% of the points, which is similar to the 1.99\% in our dataset. 
These measurements indicate that the road geometry of the two datasets is similar and encourage us to use this strategy on our dataset expecting an intrinsic error analogous to the one on KITTI.

\subsection{Viewpoint Shift Effects}
\label{subsec:exp}
Our aim is to find what are the most downgrading shifts among the different positions that we collected.
We build a training set from additional 3 videos of the \emph{base} camera with -4° pitch at 10~fps and exploit the rest of the dataset as test set at 1~fps to measure the effects of different viewpoints on MonoViT. We run our experiments using MonoViT pre-trained on KITTI with a resolution of 1024x320. To fit our images in the network, we centrally crop our frames from 720 to 704, ending up with frames of 1280x704 pixels.

We first ran a zero-shot experiment on both cameras and all the positions. Then we fine-tuned the model on the training set using  an Nvidia L4 GPU with 24GB, with a batch size of 2 for 21 epochs. For the optimization we used AdamW \cite{adamw} with an initial learning rate of $10^{-9}$ for the depth encoder and $10^{-8}$ for all the other parts of the networks, and exponential decay.

As mentioned in Section \ref{subsec:kitti}, predictions need to be scaled to metric distances for evaluation.
However, in order to highlight the effects of the shifts on the scale perceived by the model, we choose to use the same scaling factor for all positions. This factor is computed exploiting all the objects $O$ in the training set $T$, both in the zero-shot and fine-tuning experiments:
\begin{equation}
\label{eq:scale}
    s = \frac{\text{median}\left(\text{gt}(O^{k})\right)}{\text{median}\left(\text{pred}(O^{k})\right)}  \; ,  \;  \forall  O^k \in T.
\end{equation}

Table \ref{tab:results} gives the absolute relative errors of the model in the different positions, both in zero-shot and fine-tuned inference. To isolate the effect of the viewpoint, for each position, we test on footage from the same scene recorded by the two cameras simultaneously, and compute the difference in performance between \emph{shifted} (S) and \emph{base} (B) camera. The rows with the highest difference between the S and B cameras correspond to the installation positions with the strongest performance degradation.

 \begin{table}[tb]
 \begin{threeparttable}[]
   \caption{
     Absolute relative error in different camera positions
   }
   \label{tab:results}
   \centering
   \setlength{\tabcolsep}{2.75pt}
   \renewcommand{\arraystretch}{1.05}
   \begin{tabular}{@{} rrrrrrr|rrr|rrr @{}}
     \toprule
 \multicolumn{7}{c|}{\emph{Shifted} Dashcam Position}&
 \multicolumn{3}{c|}{Zero-shot}&
 \multicolumn{3}{c}{Fine-tuned}\\ 
 ID & X & Y& Z & Pitch & Roll & Yaw &B&S&S-B&B&S&S-B\\ 
 \midrule
 0&0&30&0&-4&0.0&0&18.3&18.5&0.2&17.5&17.7&0.3\\
 \midrule
 1&5&0&0&-10&0.0&0&19.3&29.2&\underline{9.8}&18.0&23.5&\underline{5.6}\\ 
 2&5&0&0&16&0.0&0&17.2&25.5&\underline{8.3}&16.9&21.1&4.2\\ 
 3&5&0&0&-4&2.5&0&18.3&21.5&3.1&15.1&16.9&1.8\\ 
 4&5&0&0&-4&5.0&0&21.0&21.2&0.2&16.9&17.6&0.7\\ 
 5&5&0&0&16&5.0&0&20.4&40.0&\textbf{19.6}&19.7&36.3&\textbf{16.5}\\ 
 \midrule
 6&-65&-5&-10&5&0.0&-5&19.5&24.1&4.7&16.6&18.7&2.1\\ 
 7&-65&-5&-10&16&0.0&-5&18.9&46.8&\textbf{27.9}&17.5&44.5&\textbf{27.0}\\ 
 \midrule
 8&-65&30&-10&5&0.0&-5&18.4&23.6&5.1&17.0&21.7&\underline{4.6}\\ 
 9&-65&30&-10&16&0.0&-5&19.0&25.4&6.4&14.8&17.9&3.1\\
   \bottomrule
   \end{tabular}
   \begin{tablenotes}
       \item B and S are the abs-rel error on the \emph{base} and \emph{shifted} cameras, respectively. S-B is the degradation computed as difference of errors. The top 2 largest degradations are in \textbf{bold}, the second 2 are \underline{underlined}.
   \end{tablenotes}
   \end{threeparttable}
 \end{table}

From the zero-shot results, we observe that the two most affected positions are 7 and 5 (in decreasing order). The combination of pitch and yaw has the most degrading effect, leading to more than doubling the error. Pitch and roll have also the effect of doubling the error of the model. Fine-tuning reduces the error in general, however the degrading effects of the 2 worst positions remain high.
In general we observe that pure rotations or translations don't affect the model as much as a combination of them, with some more impactful than others.

Given the single scale factor used for all positions, we believe that the main effect of viewpoint shifts is the distortion of the scale perceived by the model, which could not have been detected if per-image or per-position scaling was used.
In order to better investigate this phenomenon, we compute scaling factors as in (\ref{eq:scale}) on the different positions separately. On the right side of Table \ref{tab:results} we report the computed scales on the zero-shot experiment where a correlation between performance degradation and perceived scale can be seen.
In particular, we observe that higher pitch angles correspond to smaller scales perceived by the model (and therefore a bigger scaling factor necessary to align to GT). In the same way, lower pitch angles generate bigger scales perceived.
This observation seems to confirm that indeed depth models use vertical position as depth cue for objects as stated by \cite{dijk2019}.
In general this suggests that adaptation methods tackling this problem should focus on automatic scaling of models at inference time.

 \begin{table}[tb]
 \begin{threeparttable}[]
   \caption{
     Scale perceived in different camera positions
   }
   \label{tab:results_scale}
   \centering
   \setlength{\tabcolsep}{2pt}
   \renewcommand{\arraystretch}{1.05}
   \begin{tabular}{@{} rrrrrrr|rrrr @{}}
     \toprule
 \multicolumn{7}{c|}{\emph{Shifted} Dashcam Position}&
 \multicolumn{4}{c}{Zero-shot}\\ 
 ID & X & Y& Z & Pitch & Roll & Yaw & Error S-B & Scale B & Scale S & Scale S-B\\ 
 \midrule
 0&0&30&0&-4&0.0&0&0.2&31.3&30.6&-0.6\\ 
 \midrule
 1&5&0&0&-10&0.0&0&\underline{9.8}&33.7&26.9&\underline{-6.9}\\ 
 2&5&0&0&16&0.0&0&\underline{8.3}&31.6&43.2&\underline{11.6}\\ 
 3&5&0&0&-4&2.5&0&3.1&33.1&31.1&-2.1\\ 
 4&5&0&0&-4&5.0&0&0.2&36.0&33.6&-2.4\\ 
 5&5&0&0&16&5.0&0&\textbf{19.6}&33.5&52.5&\textbf{19.0}\\ 
 \midrule
 6&-65&-5&-10&5&0.0&-5&4.7&33.2&30.4&-2.7\\ 
 7&-65&-5&-10&16&0.0&-5&\textbf{27.9}&33.2&61.2&\textbf{28.0}\\
 \midrule
 8&-65&30&-10&5&0.0&-5&5.1&31.7&27.3&-4.4\\ 
 9&-65&30&-10&16&0.0&-5&6.4&34.1&39.9&5.8\\ 
   \bottomrule
   \end{tabular}
   \begin{tablenotes}
       \item Error S-B is the error degradation reported from Table \ref{tab:results} for reference. Scale B and Scale S are the scales perceived in the \emph{base} and \emph{shifted} cameras. Scale S-B is computed as Scale S - Scale B. The top 2 largest degradations are in \textbf{bold}, the second 2 are \underline{underlined}.
   \end{tablenotes}
   \end{threeparttable}
 \end{table}

\section{Conclusion}
\label{sec:conclusions}
In this paper, we addressed the underexplored problem of viewpoint shifts in monocular depth estimation for camera systems. We introduced a novel dataset collected using two dashcams in various positions and orientations. Our method leverages homography estimation and object detection as a cost-effective alternative to expensive LIDAR sensors for ground truth depth generation. We validated our approach on the KITTI dataset, demonstrating a high correlation (Spearman score of 0.97) with LIDAR measurements and measuring an absolute relative error comparable to the one using LIDAR. 

Using the MonoViT model, we quantified the effects of viewpoint shifts on depth estimation performance. Our experiments revealed that certain camera positions significantly degrade accuracy. In particular, combinations of pitch and roll and pitch and yaw have the most degrading effects. As a matter of fact, the most impactful shift induces an increased error of above 26 points in both zero-shot and fine tuning, which is more than double the error of the \emph{base} position.
Moreover, a study on inference scale of the model highlighted that camera pitch directly affects perceived scale and that, in general, performance degradation is correlated to scale distortion, suggesting that indeed automatic scaling of depth models at inference time could help mitigate the problem and thus be a promising research direction for generalization.

\bibliographystyle{IEEEtran}
\bibliography{IEEEabrv,root}
\end{document}